# Brain-like associative learning using a nanoscale non-volatile phase change synaptic device array

Sukru Burc Eryilmaz[1,*], Duygu Kuzum[2], Rakesh Jeyasingh[1], SangBum Kim[3], Matthew BrightSky[3], Chung Lam[3] and H.-S. Philip Wong[1]

[1]Department of Electrical Engineering, Stanford University, Stanford, CA, USA

[2]University of Pennsylvania, Philadelphia, PA

[3]IBM Research, T.J. Watson Research Center, Yorktown Heights, NY

Correspondence:

Sukru Burc Eryilmaz

Department of Electrical Engineering and Center for Integrated Systems,

Paul G. Allen B113X

420 Via Palou,

Stanford University, Stanford, CA 94305-4075

eryilmaz@stanford.edu

**Abstract:** Recent advances in neuroscience together with nanoscale electronic device technology have resulted in huge interests in realizing brain-like computing hardwares using emerging nanoscale memory devices as synaptic elements. Although there has been experimental work that demonstrated the operation of nanoscale synaptic element at the single device level, network level studies have been limited to simulations. In this work, we demonstrate, using experiments, array level associative learning using phase change synaptic devices connected in a grid like configuration similar to the organization of the biological brain. Implementing Hebbian learning with phase change memory cells, the synaptic grid was able to store presented patterns and recall missing patterns in an associative brain-like fashion. We found that the system is robust to device variations, and large variations in cell resistance states can be accommodated by increasing the number of training epochs. We illustrated the tradeoff between variation tolerance of the network and the overall energy consumption, and found that energy consumption is decreased significantly for lower variation tolerance.

# 1. Introduction

Historical improvements in cost and performance of CMOS technology have relied on transistor scaling for decades. However, CMOS transistor scaling has started reaching its physical as well as economic limits (Radack and Zolper, 2008). Further scaling may prevent reliable binary operation of CMOS devices. As devices are scaled down, device to device as well as cycle to cycle variations increase (Frank et al., 2001). Conventional digital logic based architectures cannot handle large variations as they are based on deterministic operation of devices; and extra circuitry aimed at mitigating these variations results in a huge overhead, increasing the cost significantly. In addition, increase in leakage current and hence the energy consumption as a result of further scaling imply that unabated scaling of transistor size is not the optimal solution for further performance increases (Frank et al., 2001). Furthermore, conventional information processing systems based on the von Neumann architecture have a performance bottleneck due to memory and processor being separated by a data channel. The increasing performance gap in the memory hierarchy between the cache and nonvolatile storage devices limits the system performance in Von Neumann architectures (Hennessy et al., 2012). Hence, in order to continue the historical performance improvements in information processing technology, different concepts and architectures need to be explored. New architectures are highly desired especially for specific applications that involve computation with a large amount of data and variables, such as large-scale sensor networks, image reconstruction tools, molecular dynamics simulations or large scale brain simulations (Borwein and Borwein, 1987).

Massive parallelism, robustness, error-tolerant nature, and energy efficiency of the human brain suggest a great source of inspiration for a non-conventional information processing paradigm which can potentially enable significant gains beyond scaling in CMOS technology and break the von Neumann bottleneck in conventional architectures (Mead, 1990; Poon and Zhou, 2011; Le et al., 2012). Synaptic electronics is an emerging field of research aiming to realize electronic systems that emulate the computational energy-efficiency and fault tolerance of the biological brain in a compact space (Kuzum et al., 2013). Since brain-inspired systems are inherently fault tolerant and based on information processing in a probabilistic fashion, they are well-suited for applications such as pattern recognition which operates on large amounts of imprecise input from the environment (Le et al., 2012). One approach to brain-like computation has been the development of software algorithms executed by supercomputers. However, since these have been executed on conventional architectures, they have not come close to the human brain in terms of performance and efficiency. For instance, IBM team has used the Blue Gene supercomputer for cortical simulations at the complexity of a cat brain (Preissl et al., 2012). Although this is a multi-core architecture, it is still nowhere close to the human brain in terms of parallelism, even though it already requires large amount of resources: 144 TB of memory and 147,456 microprocessors, and consumes 1.4 MW of power overall (as opposed to approximately 20 W consumed in biological brain in humans) (Preissl et al., 2012). Another approach is to realize brain-like parallelism in hardware instead of programming conventional systems by software. Typically, the number of synapses (connection nodes between neurons) are much larger than number of neurons in a neural network, making synaptic device the most crucial element of the system in terms of area footprint and energy consumption to realize brain-like computing systems on hardware (Drachman, 2005). CMOS implementations of smaller scale physical neural networks on a specialized hardware have been previously demonstrated (Indiveri et al., 2006). The large area occupied by CMOS synapses limits the scale of the brain-like system that can be realized with these approaches. For instance, the synaptic element in (Merolla et al., 2011) is an 8-transistor SRAM cell, with an area of 3.2 μm × 3.2 μm using a 45 nm CMOS technology. This area-inefficient synaptic element makes it impractical to scale up the system. Implementing synaptic functionality in a much more compact space, such as on the order of few tens of nanometers, would be useful to build a more compact intelligent architecture, besides potentially being more power efficient. Such a compact synaptic device is especially required when the goal is to upscale the system to the scale of human brain. In recent years, different types of emerging nanoscale non-volatile memory devices, including phase change memory (PCM) (Kuzum et al., 2011; Bichler et al., 2012; Suri et al., 2012), resistive switching memory (RRAM) (Xia et al., 2009; Yang et al., 2012; Chang et al., 2011; Seo et al., 2011; Yu et al., 2011; Yu et al., 2013) and conductive bridge memory (CBRAM) (Jo et al., (2010); Ohno et al., (2011)), have been proposed for implementing the synaptic element in a compact space. Such devices, which can be scaled to nanometer dimensions, would enable realization of highly dense synaptic arrays approaching human scale implementation of brain emulators or intelligent systems on hardware, owing to their small feature sizes. Among these different types of emerging memory devices, phase change memory has the advantage of being a more mature technology. In addition, phase change memory has excellent scalability. In fact, phase change material has shown switching behavior down to 2 nm size (Liang et al., 2012). Phase change memory arrays fabricated in 3-dimension have been demonstrated as an alternative approach for high density memory (Kinoshita et al., 2012). Functional arrays of phase change memory cells have already been demonstrated in 20 nm and other technology nodes (Kang et al., 2011; Servalli et al., 2009). Hence, it is possible to build a hybrid brain-like system using nanoscale synaptic devices using phase change memory integrated with CMOS neurons.

The main characteristic of PCM that makes it a good candidate as a synaptic device is its capability for being programmed to intermediate resistance states between high and low resistance values, or gradual programming (Kuzum et al., 2011). As illustrated by Kuzum et al., the ability to program a PCM in 1% grey-scale conductance levels enables the PCM to emulate the spike-timing-dependent plasticity (STDP) in synaptic strength in hippocampal synapses. Furthermore, the crossbar architecture used in most memory array configurations is actually analogous to grid-like connectivity of brain fibers in human brain (Wedeen et al., 2012).

The low resistance state of PCM is called the SET state and transition from the high resistance state to the low resistance state is called SET. High resistance state of PCM is called the RESET state and transition from low resistance state to the high resistance state is called RESET. Applying appropriate voltage pulses create intermediate resistance states between the fully SET state and the fully RESET state in a phase change memory device (Kuzum et al., 2011). This is similar to gradual weight change in biological synapses, where the synaptic weight is modified in accordance with relative arrival timing of the spikes from pre and post-neurons. This is called spike timing dependent plasticity (STDP), and is thought to be one of the fundamental learning rules in hippocampal synapses (Bi and Poo, 1998). Using this property of phase change devices as well as similar characteristics of other emerging memory devices mentioned above, network level learning studies have been done (Bichler et al., 2012; Yu et al., 2013; Kaneko et al., 2013; Pershin and Di Ventra, 2010; Pershin and Di Ventra, 2011; Alibart et al., 2013). However, many of these works studying nanoscale synaptic devices on network level have been limited to simulations, and experimental works either have used few number of synapses or lack a thorough analysis of important network parameters (Kaneko et al., 2013; Pershin and Di Ventra, 2010; Alibart et al., 2013). Recently, we presented preliminary findings of hardware demonstration of a synaptic grid using phase change memory devices as synaptic connections (Eryilmaz et al., 2013). In this work, we present a detailed description of the algorithm and signaling scheme used, and additionally present a thorough analysis of the tradeoff between the power consumption, the number of iteration required, and the device resistance variation. We experimentally study the effects of resistance variation on learning performance in the system level. We find that larger variations can be tolerated by increasing the number of learning epochs, but this comes with increased overall energy consumption, resulting in a trade-off between variation tolerance, energy consumption, and speed of the network.

## 2. Phase Change Memory Cell Array for Synaptic Operation

Phase Change Memory (PCM) cells used in the experiment are mushroom type cells, which means the heater material, bottom electrode (BE), phase change material, and the top electrode (TE) are stacked on top of each other, respectively (Wong et al., 2010). The 10-by-10 memory array used in the experiments consists of 100 memory cells. These cells are connected in a crossbar fashion as illustrated in Figure 1(a). Each memory cell consist of a PCM element in series with a selection transistor. The circuit schematic of a memory cell is shown in Figure 1(a), and a cross section of a memory cell is shown in Figure 1(b), together with the optical microscope image of the memory chip used. The cells can be accessed through bitline (BL) and wordline (WL) nodes. Each wordline is connected to the gates of selection transistors of 10 memory cells, and each bitline is connected to the top electrode of the PCM element of 10 memory cells. Overall, there are 10 WL and 10 BL nodes in the array. Note that the bottom electrode of a PCM element within a cell is connected to the selection transistor of that cell. Each cell is associated with a unique (WL, BL) pair, hence each cell can be accessed by applying bias to the corresponding BL and WL nodes, as shown in Figure 1(a). The device fabrication as well as retention and endurance characteristics of memory cells in the array are given in detail elsewhere (Breitwisch et al., 2007).

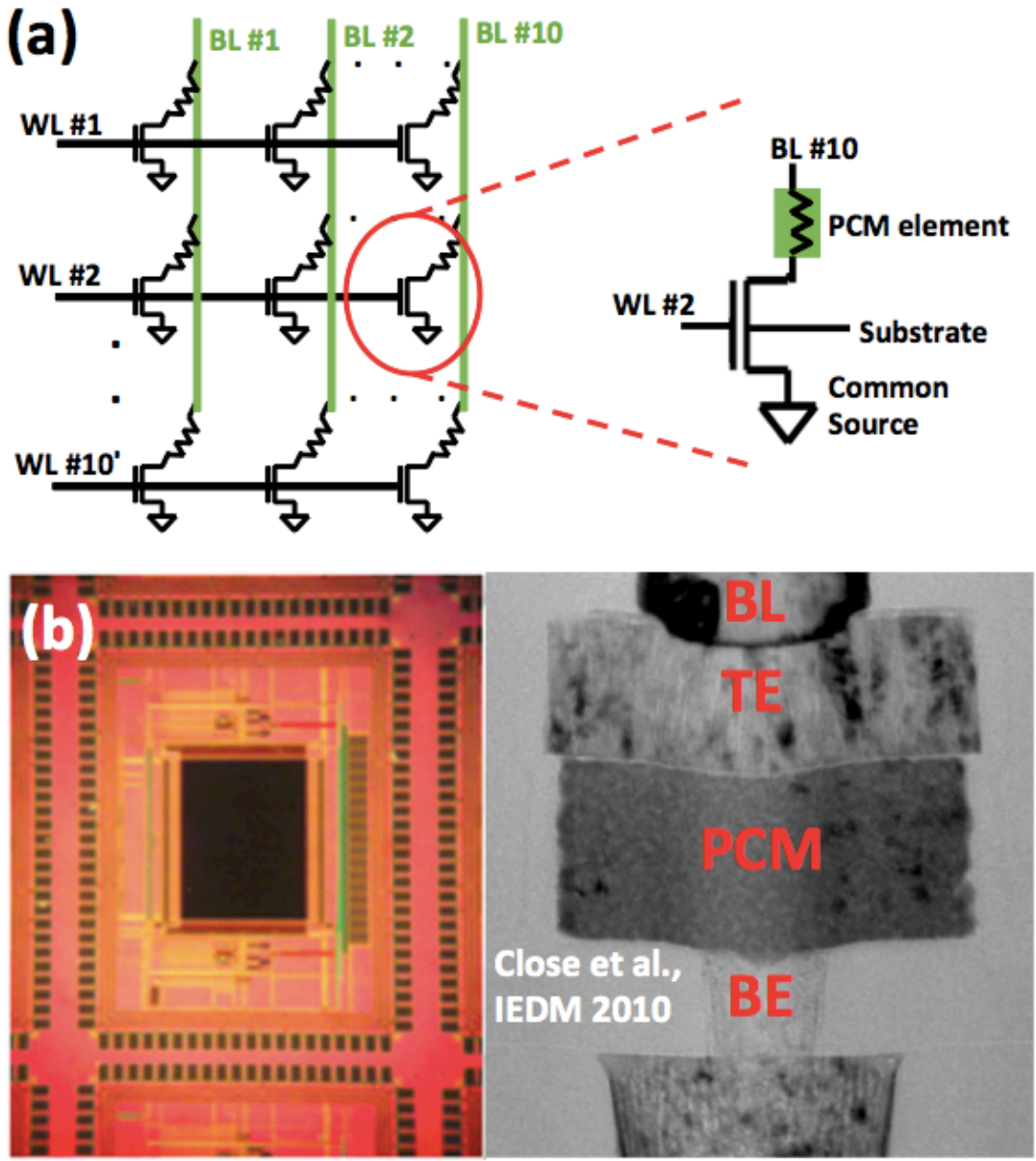

**Figure 1.** (a) Schematic of 10x10 phase change memory (PCM) cell array is shown on the left. Resistances connected in series with the selection transistors represent PCM element. The figure on the right shows the complete schematic of a single memory cell. This particular cell can be accessed by applying appropriate biases at WL #2 and BL #10. Substrate and common source terminals are grounded during the experiment. (b) Optical microscope image of memory cell array located on the memory chip is shown on the left. TEM image of a single memory cell is appended to the right hand side. Mushroom type cell structure can be seen by observing the bitline (BL), top electrode (TE), phase change material (PCM) and bottom electrode (BE) stack. TEM image is reprinted with permission from Close et al., 2010. Copyright 2010 IEEE. TEM image is a representative figure for 90 nm node mushroom PCM cell, and PCM cells in the array in this paper are 180 nm node with the same device structure.

SET programming of a memory cell is achieved by applying a long (from a few hundred ns to few µs) current pulse through the PCM element to crystallize the phase change material in the PCM via Joule heating. In a gradual SET programming, depending on the amplitude of the current pulse, resistance of the PCM reduces for a certain amount, rather than going directly into the lowest resistance (fully SET) state (see Figure 2(d)). RESET (high resistance) programming is achieved by amorphizing the phase change material of the memory cell by applying a larger current pulse with a very sharp (2-10 ns fall time) falling edge. A large amplitude of current pulse results in melting of PCM material through Joule heating, the sharp falling edge quenches the cell, without allowing time for the phase change material to go into the more stable crystalline state, leaving it in the amorphous state. The amount of resistance increase for gradual RESET can be controlled either by changing the falling edge width of the current pulse or by changing the current pulse amplitude (Kang et al., 2008; Mantegazza et al., 2010). Typical DC switching characteristics of a single device arbitrarily chosen from an array are shown in Figure 2(a). For DC switching characterization, 3.3 V is applied at WL of a single cell and BL node is swept from 0 V up to the switching threshold. The measurement result in Figure 2(a) shows that switching threshold for one of the cells in a fully RESET state is around 0.8 V, and the current when switching occurs is 2 µA. Note that these values can vary across the memory array due to device to device variation. Set and reset pulses with amplitudes of 1 V and 1.5 V and with (50 ns/300 ns/1 µs) and (20 ns/50 ns/5 ns) rise/width/fall time is applied at WL node, while BL node is held at 3.3 V during characterization of pulse switching in the memory cells. Pulse switching characteristics are shown in Figure 2(b). This data is obtained by applying SET pulses for pulse #1,3,5… and RESET pulses for pulse #2,4,6…. The same SET and RESET pulses are used for array level binary resistance characterization shown in Figure 2(c). RESET resistance is distributed around 3M ohms and SET resistance is distributed around 10k ohms. For synaptic operation, gradual resistance change characteristics of memory cells are utilized. Specifically, our system utilizes gradual SET programmability of memory cells. To characterize gradual resistance change from the RESET state to the partially SET state, we apply once a 1.1 V RESET pulse and then 9 SET pulses with 0.85 V amplitude. Gradual resistance change characteristics from RESET to SET for a single cell is shown in Figure 2(d) for a few cycles of gradual SET characterization. This gives us around 9 resistance levels between low and high resistance state. Although the energy consumption for gradual SET is lower than gradual RESET, variability is larger for gradual SET since gradual SET is probabilistic in nature (Braga et al., 2011). The reason behind this is the intrinsic stochasticity of the nucleation of crystalline clusters during gradual SET operation. The cycle-to-cycle variability is also observed in Figure 2(d) The same resistance levels are not accurately repeatable from cycle to cycle. Due to variability in gradual resistance change, multi-level-cell (MLC) memory applications use a write-and-verify technique since the data storage applications require deterministic binary resistance levels (Kang et al., 2008). However, massively-parallel brain-like architectures can tolerate such variations and do not require the use of write-and-verify that is needed to achieve an accurate resistance level. Hence, the variations observed in Figure 2(d) do not pose a problem for our purposes.

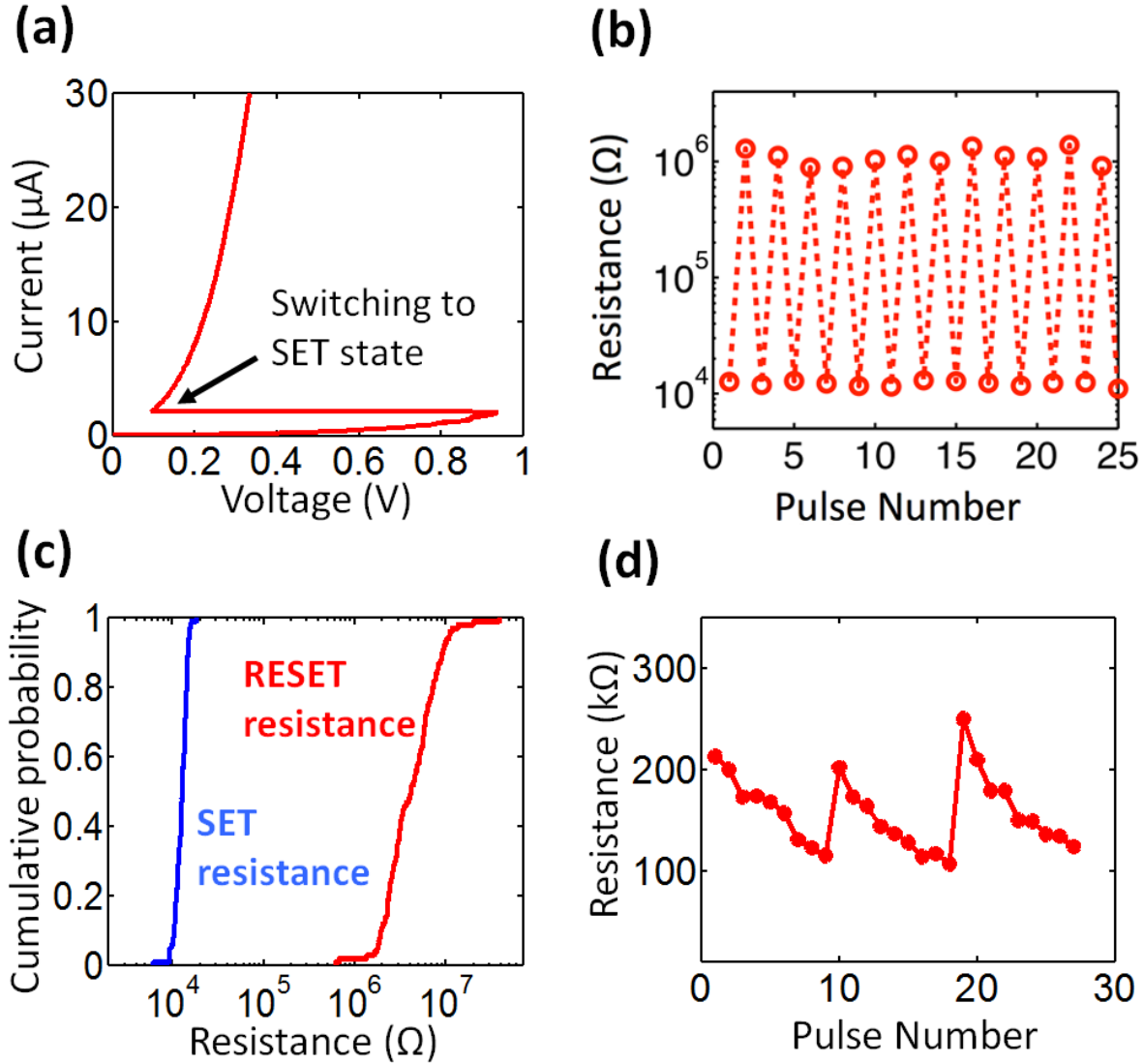

**Figure 2.** Electrical characterization of memory cells. (a) shows the DC switching characteristics of a single memory cell arbitrarily selected from the array. Switching behavior can be observed when there is 2 µA of current through the memory cell. Binary switching cycles are shown in (b). SET pulse is applied at odd numbers of measurement (pulse #1, 3, …) and RESET pulse is applied at even numbers of measurement. The plot shows the measured resistance of the memory cell right after the programming pulse is applied. Array level binary resistance distribution is shown in (c). Resistance window for binary operation is larger than 10k. Gradual resistance change in a single cell is shown in (d). This plot is obtained by applying gradual SET pulses right after the cell is abruptly programmed to RESET state. The plot shows 3 cycles of this measurement.

## 3. Array Level Learning

A fully-connected recurrent Hopfield network is employed for the learning experiments (Figure 3(a)) (Hertz et al., 1991). The Hopfield network consists of 100 synaptic devices and 10 recurrently connected neurons, as shown in Figure 3(a). It is worth noting that in this architecture, all neurons are both input and output neurons. Integrate-and-fire neurons are implemented by computer control and memory cells serve as synaptic devices between neurons. Figure 3(a) illustrates how the network is constructed using the memory cell array. The input terminal of the i-th neuron is connected to BL #i, and output terminal of the i-th neuron is connected to WL #i, where i=1,2,…,10, i.e., neuron #1 input and output is connected to BL #1 and WL #1, respectively, and neuron #2 input and output is connected to BL #2 and WL #2, respectively, etc. (Figure 3(a)). Before the experiment, all synapses are programmed to the RESET state. A learning experiment consists of epochs during which synaptic weights are updated depending on firing neurons. After training, the pattern is presented again but with an incorrect pixel this time, and the incorrect pixel is expected to be recalled in the recall phase after training is performed (Figure 3(b)). A complete pattern is presented during the training phase of an epoch, and an incomplete pattern with an incorrectly OFF pixel is presented during the recall phase. All patterns consist of 10 pixels, and each neuron is associated with a pixel. This mapping between pixels and neurons is shown in Figure 3(c) for two different patterns considered in this work. Figure 3(b) shows the pulsing scheme for firing and non-firing neurons in both update and recall phases. When a pattern is presented during a training phase, the neurons associated with ON (red pixels in Figure 3(c)) pixels are externally stimulated, hence they fire. As can be seen in Figure 3(d), when a neuron spikes during the training phase, it applies programming pulses at its input (corresponding BL) and output (WL). This results in gradual SET programming of the synaptic device between those two firing neurons. For instance, when neuron 1 and neuron 2 fire, programming pulses are applied at WL1, WL2, BL1 and BL2, as defined in the pulsing scheme in Figure 3(b). These pulses will result in a current going through PCM elements and hence gradual SET programming of memory cells that connect neuron 1 and neuron 1 (see Figure 3(d)). After training, the recall phase begins. During the recall phase, a pattern with an incorrectly OFF pixel is presented (Figure 3(e)). Again, the neurons associated with ON pixels during recall phase fire, and appropriate pulses are applied at the input and output of neurons as shown in the pulsing scheme in Figure 3(b). Neurons associated with OFF pixels during recall phase do not fire. Note that there is a low amplitude pulse applied at the input of non-firing neurons during recall phase. This voltage pulse, together with the large amplitude voltage pulse applied at the firing neurons' output during recall phase, create an input current feeding into non-firing neurons. The amplitude of this current through a non-firing neuron is determined by the resistance values of synaptic connections between that neuron and the firing neurons. This input current of non-firing neurons during recall phase is analogous to membrane potential of biological neurons. In biological neurons, the postsynaptic current feeding into a neuron accumulates charge on capacitive membrane, forming a membrane potential. Typically, this is modeled by a time constant that is determined by membrane capacitance. In this experiment, neurons fire simultaneously during the recall phase, while at the same time the input current through the non-firing neurons is measured. Since the delays and timing properties of the neurons are not included in the neuron model, the membrane capacitance is not included in neurons. Hence, input current through a neuron is actually equivalent to membrane potential in our experiments. Note that in this paper, we will use the terms input current and membrane voltage interchangeably, due to the reasons explained above. The input current into a non-firing neuron during recall phase can be written as follows:

$$I_i = V_{read} \sum_{j \in F} \frac{1}{R_{ij}} \qquad (1)$$

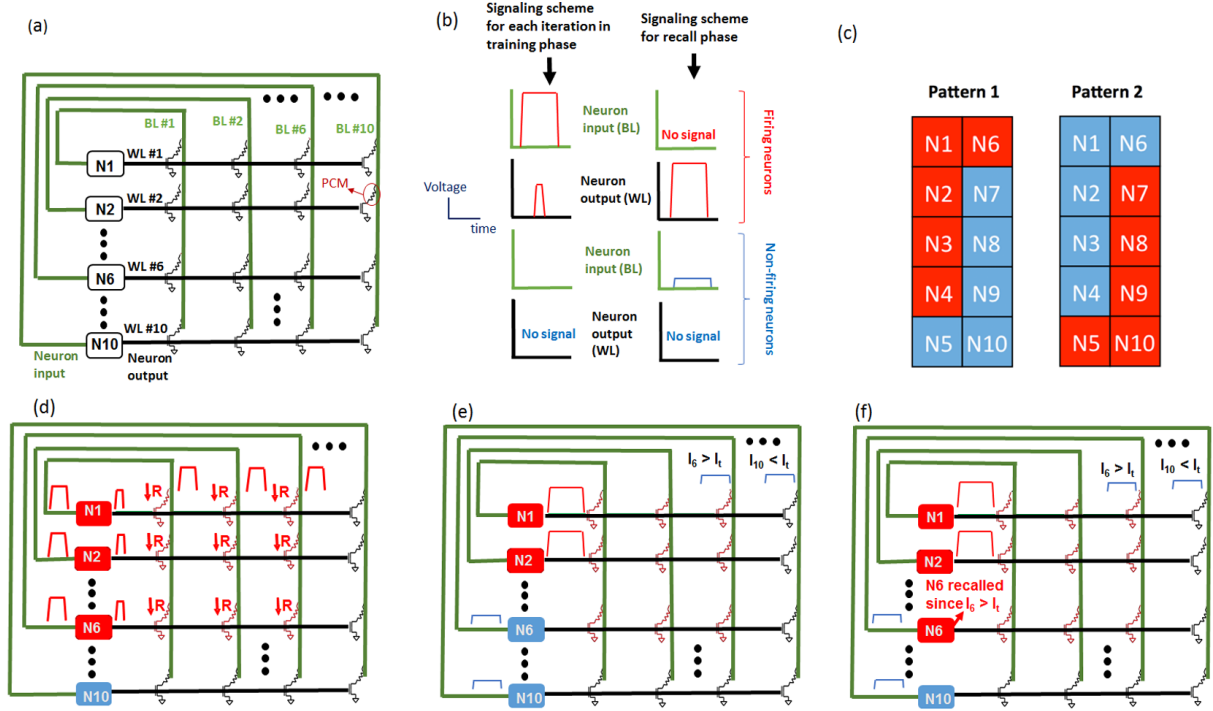

**Figure 3.** Neural network realized and how it is implemented with the memory array is explained. (a) shows the recurrently connected Hopfield network implemented in the learning experiment. Pulsing scheme during training as well as recall is shown in (b). We train the network with two patterns as shown in (c), where red pixels correspond to ON and blue pixels correspond to OFF. Numbers in pixels correspond to the neuron number associated with that pixel. During update phase shown in (d), the resistance of synaptic elements connected to non-firing neurons do not change, since no pulse is applied at the WL node of non-firing neurons during update phase. The synaptic connections between firing neurons, however, are programmed by the pulses applied at the BL and WL of the corresponding memory cell. The pulse characteristics are predetermined for gradual SET programming of the memory cell, hence the resistance is reduced with an amount and the connection gets stronger. (e) During the read phase, a small amplitude voltage applied at the BL node of non-firing neurons sense the total current due to the synapses of that neuron connected to firing neurons, since a pulse applied at the output of the firing neurons turns the selection transistor on simultaneously. (f) In this example, during the recall phase, N1, N2, N3 and N4 are presented with N6 OFF (not firing), but N6 is recalled since the input current of N6 is larger than the threshold.

In eq. 1, $I_i$ is the input current into the i$^{th}$ neuron where it is a non-firing neuron, F is the set of indices of firing neurons, $R_{ij}$ is the resistance of synaptic element between bitline i and wordline j, and $V_{read}$ is the read voltage at the input of non-firing neurons during recall phase (see Figure 3(b)), which is 0.1 V in our experiments. As Figure 3(b) shows, if a neuron is not associated with an OFF pixel at the beginning of the recall phase, it fires, and the reading voltage $V_{read}$ at its input is 0, making its input 0.

If the input current through a non-firing neuron exceeds a threshold during the recall phase, then the neuron associated with the pixel fires, the complete pattern is recalled (Figure 3(f)). The membrane potential of neurons is set to 0 at the beginning of each epoch, hence it does not transfer to the next epoch. We define "missing pixel" as the pixel that is ON in the correct pattern used for training, but OFF in the input pattern during recall phase. Note that the pixel missing from the pattern in recall phase still fires in update phase during training, SET programming the corresponding memory cells between this neuron and other firing neurons. This results in a decrease in the resistance values between this missing pixel's neuron and other firing neurons (ON pixels) as shown in Figure 3(d), increasing the input current of the missing pixel's neuron during the recall phase (Figure 3(f)). Hence, recall is expected to occur after a few epochs, at which point the membrane potential exceeds a pre-determined threshold. This learning scheme is a form of Hebbian learning, since the weights of synaptic connections between coactive neurons during training phase get stronger, due to reduced resistances of these synaptic connections. The time window that defines the firing of two neurons as being coactive is determined by the width of the pulse applied at the input of firing neurons during update phase, shown in Figure 3(b). This time window is 100 $\mu$s in our experiments. As an illustration of the aforementioned learning process, two simple 10-pixel patterns are chosen to be learned. The two patterns of 10 pixels are shown in Figure 3(c). The network is first trained with pattern 1 (on the left in Figure 3(c)), and then pattern 2 (on the right in Figure 3(c)). During training with pattern 1, until the pattern is recalled, the complete pattern is presented in training phase and the pattern with pixel 6 missing is presented during recall phase. After pattern 1 is recalled, the same procedure is performed for pattern 2, this time with pixel 5 missing in the recall phases of epochs. This experiment is performed for 4 cases, each corresponding to different initial resistance variations across the array. Initial variation here refers to the variation after all cells are programmed to RESET before learning experiment begins. Different initial variation values are obtained by individually programming the memory cells in different arrays. The evolution of synaptic weights is shown in Figure 4 during the experiment for the case where the initial variation is 60%. Note that the synaptic weight map in Figure 4 shows the normalized synaptic weights of each synaptic device. Each data point in this map shows the resistance of the synaptic device after the corresponding epoch divided by the initial RESET resistance (right before the experiment when all devices are RESET programmed as explained above) of that device. Hence the map does not include the variations of initial RESET state resistances across the array. The variation study is explained in the next section. As can be seen in Figure. 4, after feeding each input pattern into the network, synapses between the ON neurons gradually get stronger (resistance decreases); after 11 epochs, patterns are recalled. The overall energy consumed in synaptic devices during this experiment is 52.8 nJ. This energy does not include the energy consumed in the neurons and the wires, and is the energy consumed by the synaptic devices during training and recalling of pattern 1. Our measurements indicate that roughly 10% of this energy is consumed in phase change material, while around 90% is consumed in selection devices in our experiment. Note that the number of epochs and the overall energy consumed strongly depends on the choice for the threshold membrane potential of neurons. If threshold membrane potential is kept low, the number of epochs would be reduced, but a wrong pixel might fire (hence turn on) in the output of recall phase due to variations, hence recalling a wrong pattern. This is explained in detail in the next section.

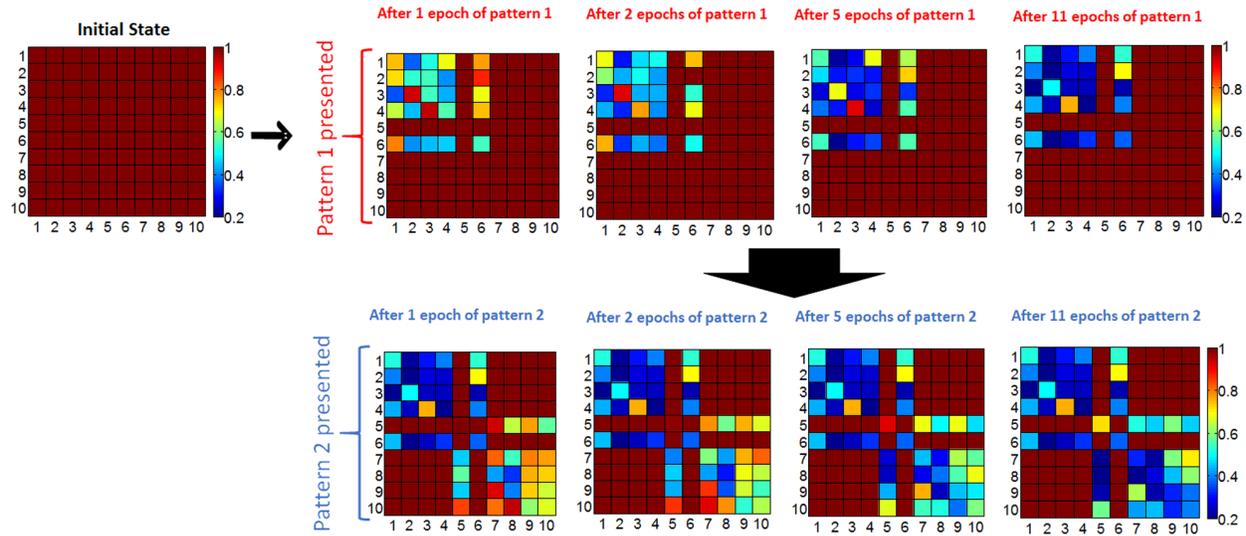

**Figure 4.** Evolution of normalized resistance of synaptic devices is shown, for the 60% initial variation case. All normalized resistances are 1 initially since the normalized resistance map shows the current resistance of a synaptic device divided by its initial resistance. Note that the row and column numbers corresponds to BL and WL that connect the synaptic devices. For instance, the data shown in row #3 and column #6 is the normalized resistance of the memory cell that can be accessed by BL #6 and WL #3. First, pattern 1 is presented to the network. For pattern 1, ON neurons for the complete pattern during update phase are N1, N2, N3, N4, N6, and, and for the recall phase N6 is OFF and expected to be recalled (i.e., expected to fire) after training with a certain number of epochs. The gradual decrease in the normalized resistance of synaptic connections between firing neurons during the update phase can be observed. After 11 epochs, when recall phase is performed, OFF pixel #6 (neuron #6) is recalled (meaning neuron #6 fires in recall phase) , and then pattern 2 is presented for training. For pattern 2, the complete pattern is represented by N5, N7, N8, N9, N10; and N5 is missing in the recall phase.

## 4. Effect of Variation on Learning Performance

Figure 5(a) shows the actual resistance map of synaptic connections after 11 epochs for the experiment above, along with the resistance distribution (on the left in Figure 5) when all the cells are in the RESET state before the experiment. As the synaptic connections evolve during training for two patterns, synapses between coactive neurons get stronger. Actual resistance maps in Figure 5 also illustrate the resistance variation across the array when all cells are in RESET state before training. In our experiment, the neuron firing threshold is the important parameter that can be tuned to tolerate the variation. This threshold value has to be large enough so that a wrong pixel will not turn on in recall phase, but low enough to guarantee that the overall energy consumed is minimal and the missing pixel will actually turn on in recall phase, hence recalling the original pattern. To this end, the firing threshold of neurons is selected as follows:

$$I_{thr} = C \cdot \max_{N,i} \left( V_{read} \sum_{j \in N} \frac{1}{R_{ij}} \right) \qquad (2)$$

In eq. 2, N is constrained to be a 4-element subset of the set {1,2,3,…9,10}, and $R_{ij}$ is the initial RESET resistance of the memory cell defined by bitline i and wordline j, and $V_{read}$ is defined as in eq. 1. This equation means that the threshold current is a constant C times the largest input current that a neuron can possibly have in the recall phase, given the resistance values for each cell. The reason for considering 4-element subsets is because we are assuming 4 pixels are ON in the input during recall phase, and we want to make sure that the threshold is large enough to avoid firing of a neuron during recall phase when it is actually not ON in the true pattern. In its current form, this scheme might not be successful when different number of pixels are missing, for example, when three pixels are ON in recall phase while 5 pixels are ON in the actual pattern. This generalization can be made by allowing negative weights; equivalently using 2-PCM synapse suggested in (Bichler et al., 2012), or adaptive threshold method suggested in (Hertz et al., 1991). The requirement that C>1 guarantees that during the training, the wrong pixel will not be recalled at any epoch. This is because the resistance of the synaptic connections between an arbitrary OFF pixel in the original pattern and other neurons do not decrease, as the OFF pixels do not fire during training. We choose C=2 for our experiments. Choosing C=2 also allows us, without requiring negative synaptic weights, to generalize recall to some extent for inputs with incorrectly ON pixels, in addition to incorrectly OFF pixels as given in our example. This idea is similar to adaptive threshold method in (Hertz et al., 1991), where instead of using negative weights, neuron threshold is increased while keeping the weights positive. Observe that as the variation increases, the low-resistance tail of the initial RESET resistance distribution (leftmost histograms in Figure 5(a)-(d)) extends towards lower resistance values. This results in a decrease in minimum resistance values, as can be seen in histograms in Figure 5(a)-(d). Hence, maximum neuron input current with 4 neurons firing increases. This increases the max term in eq. 2, hence a higher number of epochs is needed to recall the missing pixel for larger variation. The resistance maps for other variation cases are shown in Figure 5(b)-(d). We can see that as initial variation reduces, the same number of epochs yields a more pronounced overall difference between the weights that get stronger versus the weights that do not change, as illustrated in Figure 5. The evolution of the membrane potential with the number of epochs for different variation cases are shown in Figure 6. While it takes 11 epochs to recall a pattern when there is 60% initial variation, only one epoch is sufficient in our case when the initial resistance variation is 9%. It is worth mentioning that we have negligible variation in read voltage during our experiment, since the reading of memory cell resistances is performed with electronic equipment.

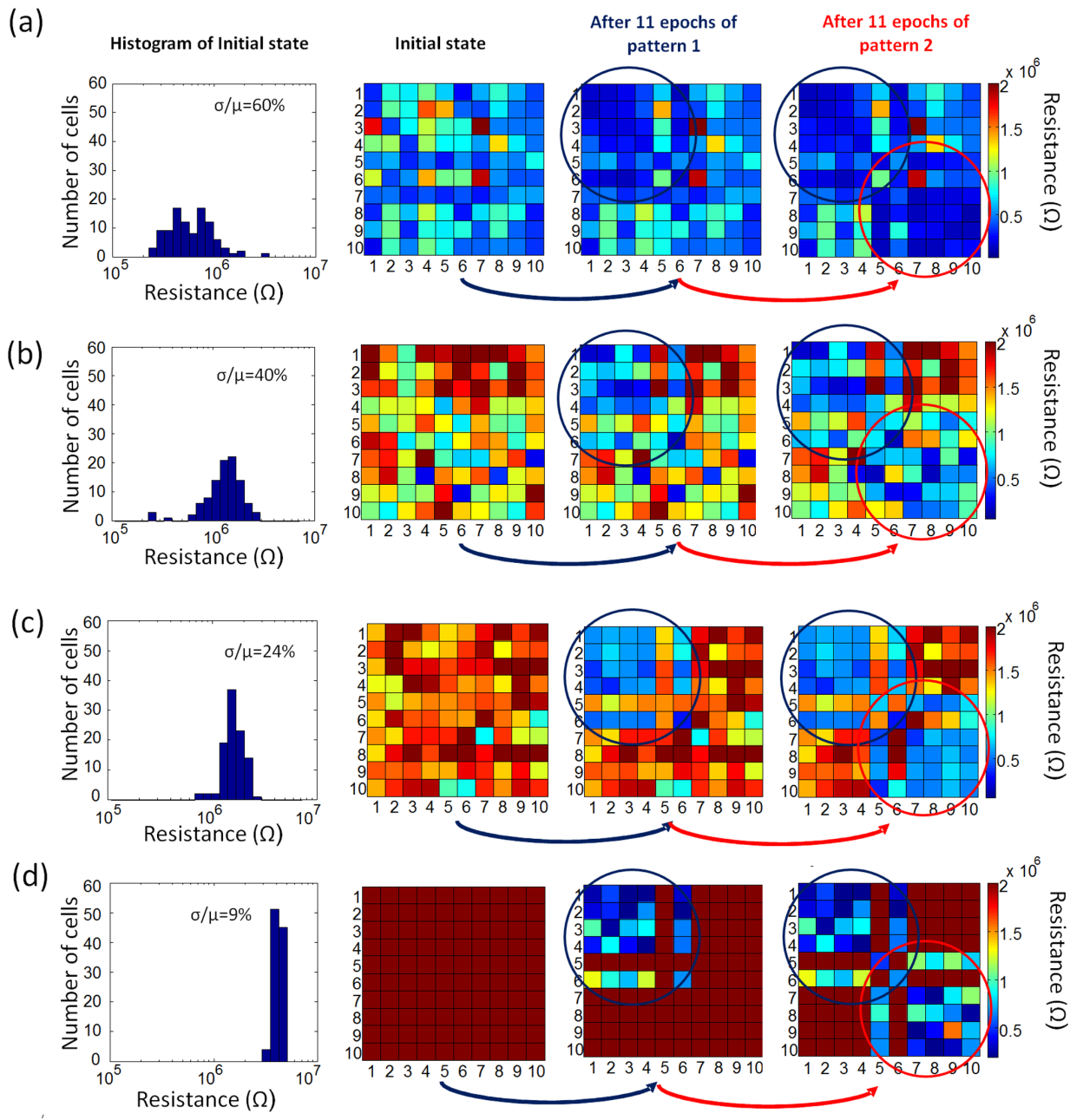

**Figure 5.** Evolution of actual resistance of synaptic devices is shown for four different experiments. The representation of synaptic devices in these resistance maps are the same as in fig. 4, but this time the resistance values are not normalized. The variations across the memory cell arrays are apparent here. Synaptic devices between firing neurons during training get stronger (i.e., are driven to lower resistance values). As the initial variation reduces, the difference in resistance values between potentiated synapses and the synapses that remain unchanged becomes more pronounced.

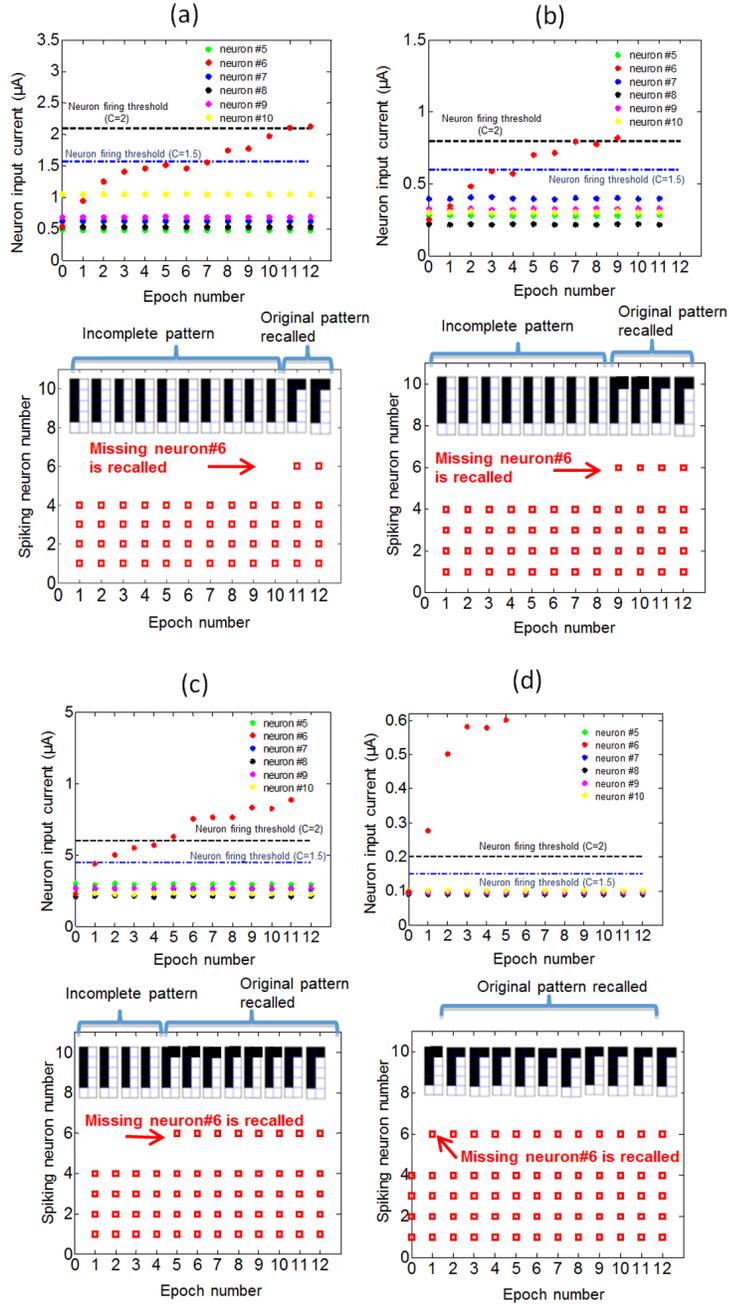

**Figure 6.** Recall of the missing pixel for training with pattern 1 for four different initial variation cases, (a) 60%, (b) 40%, (c) 24% and (d) 9%, are shown. For each case, top figures show what the input current of neurons that do not fire would be if the recall is performed after the corresponding number of epochs, and bottom figures show the neurons that fires if the recall was performed after the corresponding number of epochs for C=2 (see the text for details about parameter C). Different threshold levels for C=1.5 and C=2 cases are shown in the top figures. When the input current exceeds the threshold after a certain number of epochs, the missing pixel N6 fires. For C=2, the number of epochs after which N6 fires in each case is 11 (60% variation), 9 (40% variation), 5 (24% variation) and 1 (9% variation).

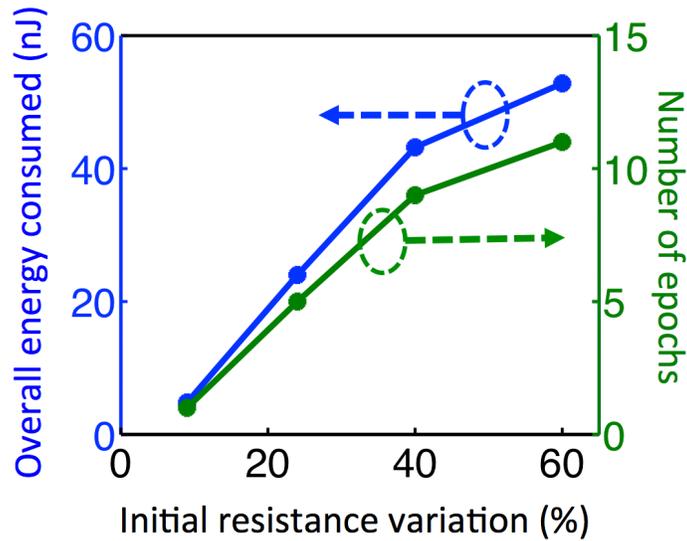

**Figure 7.** The same experiment is repeated for different initial variation cases. In order to guide the eye, dashed arrows and circles indicate which curves correspond to which axis. For four different initial variation cases, the plot shows the total number of epochs required for training as well as overall energy consumed in the synaptic devices during training and recall phases for pattern 1. As the variation increases, larger firing threshold is required for neurons. This increases the number of epochs and energy consumption required for training.

When this synaptic grid is integrated with actual CMOS neurons, however, it is expected to have some variation in read voltage, which results in variation in the input current of neurons. This variation in input current might cause variations in the number of epochs needed for training. We can observe from Figure 6 that while it takes 3% input current variation (hence read voltage variation) to change the number of epochs needed for 60% variation case (Figure 6(a)), it takes 40% variation in read voltage to change the number of training epochs required for 9% initial variation case (Figure 6(d)). This is because as the number of epochs increases, resistances of programmed synapses begin to converge to low resistance values. To minimize the effect of read voltage variation, properties of synaptic device as well as pulsing scheme during training should be carefully chosen, considering the read voltage variation of CMOS neuron circuit. The increase in the required number of epochs to recall the pattern results in a higher overall energy consumption. Overall energy consumption for 9% initial resistance variation case is 4.8 nJ, whereas it is 52.8 nJ for 60% initial variation case. Figure 7 illustrates the dependence of energy consumption and number of epochs needed on initial resistance variation. As can be seen in Figure 7, there is a clear reduction in the overall energy consumption as initial resistance variation goes down. Note that these energy values represent only the energy consumed in the synaptic devices for training and recall phases for pattern 1. They do not include the energy consumed in the wires or the neurons. Energy consumption in the wires can be a substantial part of the overall energy consumption for a large array (Kuzum et al., 2012). It is also worth noting that since the time scale between the epochs in these experiments is on the order of seconds, we did not observe any effects of drift in our measurements, which would require a timescale of μs or ms to observe (Karpov et al., 2007).

## 5. Conclusion

We report brain-like learning in hardware using a crossbar array of phase change synaptic devices. We demonstrated in hardware experiments that synaptic network can implement robust pattern recognition through brain-like learning. Test patterns were shown to be stored and recalled associatively via Hebbian plasticity in a manner similar to the biological brain. Increasing the number of training epochs provides a better tolerance for initial resistance variations, at the cost of increased energy consumption. Demonstration of robust brain-inspired learning in a small-scale synaptic array is a significant milestone towards building large-scale computation systems with brain-level computational efficiency.

## 6. Methods

The memory cell array was probed using a 25x1 probe card which is connected to a switch matrix consisting of two cards, each providing a 4x12 matrix (see Figure 8, supplementary figure). The probe card contacts 25 pads on the wafer that has the memory arrays. These 25 pads consist of 10 bitlines, 10 wordlines, 1 common source terminal, 1 substrate terminal, and 3 floating terminals. Switch matrix is connected to Agilent 4156C semiconductor analyzer to perform DC measurements and Agilent 81110 pulse generator for pulse measurements. All these equipment is controlled by a Labview program on a separate computer. This program allows us to switch between cells on the array automatically and applying custom signals from semiconductor analyzer or the pulse generator to the desired cell. In all the measurements, resistance of the memory cell is measured by applying 0.1 V read voltage at the bitline and 3.3 V at the wordline. The current (I) through the cell is measured and resistance is obtained by R = 0.1 V/I. DC switching measurement in Figure 2a is obtained from an arbitrarily selected cell on the array. For this particular measurement, current through the device is swept. For binary switching measurement in Figure 2b, alternating SET pulses (1 V amplitude, 50 ns/300 ns/1μs rise/width/fall time) and RESET pulses (1.5 V amplitude, 5 ns/50 ns/5 ns rise/width/fall time) are applied by pulse generator. For the measurement in Figure 2(c), the same SET and RESET pulses are applied at each 100 cells in an array. The gradual SET characteristics in Figure 2(d) is obtained by applying 1.1 V RESET pulse once and then 0.85 V gradual SET pulse 9 times. This cycle is repeated for a few times to obtain the result in Figure 2(d). During learning experiment, the initial RESET programming of the cells before learning experiment starts was done by applying a RESET pulse (1.5 V amplitude, 5 ns/50 ns/5 ns rise/width/fall time) at every cell within the array. The energy consumed during gradual SET programming of synaptic connections in update phases is extracted by measuring the current through the devices during programming. Fraction of energy consumed in phase change material and in selection transistor is extracted by measuring individual transistor characteristics separately, as well as by current sweep measurements in PCM cells.

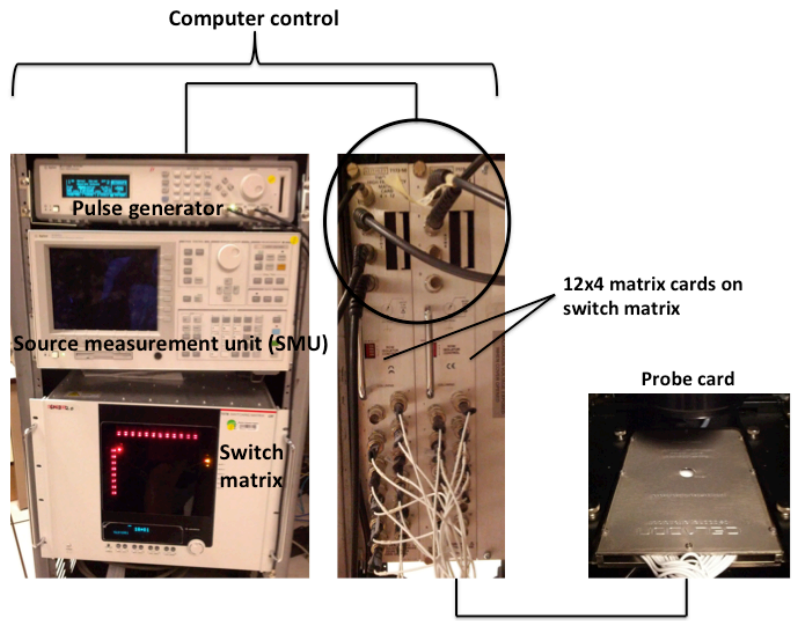

**Figure 8(Supplementary Figure).** Measurement setup used in experiments. Probe card that directly probes pads on memory chip are connected to switch matrix. Setup is controlled by computer program.

**Acknowledgements**

This work is supported in part by Systems on Nanoscale Information Fabrics (SONIC) Center, one of six centers of Semiconductor Technology Advanced Research Network (STARnet), a Semiconductor Research Corporation (SRC) program sponsored by Microelectronics Advanced

Research Corporation (MARCO) and Defense Advanced Research Projects Agency (DARPA), the NSF Expeditions in Computing (award 1317470), and the member companies of the Stanford Non-Volatile Memory Technology Research Initiative (NMTRI).

**Conflict of Interest Statement:**

The authors declare that the research was conducted in the absence of any commercial or financial relationships that could be construed as a potential conflict of interest.